\documentclass{article}
\usepackage{spconf,amsmath,graphicx}
\usepackage{multirow}

\begin{document}
\title{Learning Dense Correspondence from Synthetic Environments}
%
%
%
\name{\centering Mithun Lal \sthanks{Corresponding Author. Mail: mithun.lal@hdr.qut.edu.au} $^{,1,2}$, Anthony Paproki $^{1,2}$ , Nariman Habili $^{2}$, Lars Petersson $^{2}$, Olivier Salvado $^{1,2}$, Clinton Fookes $^{1}$ }

 \address{$^{1}$ Queensland University of Technology, $^{2}$ CSIRO, Australia}
%

%
\maketitle
\begin{abstract}
Estimation of human shape and pose from a single image is a challenging task. It is an even more difficult problem to map the identified human shape onto a 3D human model. Existing methods map manually labelled human pixels in real 2D images onto the 3D surface, which is prone to human error, and the sparsity of available annotated data often leads to sub-optimal results. We propose to solve the problem of data scarcity by training 2D-3D human mapping algorithms using automatically generated synthetic data for which exact and dense 2D-3D correspondence is known. Such a learning strategy using synthetic environments has a high generalisation potential towards real-world data. Using different camera parameter variations, background and lighting settings, we created precise ground truth data that constitutes a wider distribution. We evaluate the performance of models trained on synthetic using the COCO dataset and validation framework. Results show that training 2D-3D mapping network models on synthetic data is a viable alternative to using real data.

\end{abstract}


\section{Introduction}
\label{sec:intro}
Estimating human shape and pose from a single 2D image and establishing correspondence with a 3D human surface is a hard problem. Challenges from the lack of 3D information in 2D images range from depth ambiguity to occlusion or imaging conditions (e.g lighting) as well as the difficulty and tediousness of creating (sparse) ground truth data and overall natural variability in human pose and shape. It is a primary task for many applications including virtual reality, human action recognition, augmented reality and human-robot interaction \cite{varol2021synthetic, zherdev2021producing}. 2D-3D human mapping involves other difficult problems including object detection, pose estimation and semantic segmentation as auxiliary tasks. Patel et al. \cite{patel2021agora} introduced a synthetic human dataset for regression analysis containing 3D human scans placed in scenes that performed well on fine tuning the SMPL (Skinned Multi-Person Linear) oPtimization IN the loop (SPIN)\cite{kolotouros2019learning} model that fits a human shape model to humans in 2D images. \cite{fabbri2021motsynth} introduced another synthetic dataset that improves pedestrian detection and tracking. Thus, the adaptability and reliability of synthetic data over real data are increasing over time.

There is a substantial body of work trying to solve the problem of mapping a 2D image to a 3D model with dense point correspondence for inanimate objects \cite{grewe2016fully, taylor2012vitruvian, zhou2016learning}. For humans, 3D representations, parametric surface models such as SMPL \cite{loper2015smpl} or Adam model \cite{joo2018total} are often used to create controllable 3D surface deformations. Pixel aligned implicit functions have been used for high resolution clothed human digitization by transferring the human geometry onto a 3D model based on a single image. The method is non-parametric, which decreases the bias towards training data \cite{saito2019pifu}, however, when complex motions are involved, parametric models like SMPL can be constrained more easily and provide more temporally consistent and realistic animations.


DenseReg \cite{alp2017densereg} uses Convolutional Neural Networks (CNNs) to establish dense correspondence between 2D human face images and 3D models. Challenges involved a high structural complexity and pose variability. A 24 part IUV (Texture coordinate space) regression to achieve dense correspondence between image pixels and surface points was introduced in \cite{guler2018densepose}. A cascaded extension of Densepose-RCNN and an in-painting network were used to interpolate a dense supervision from sparse supervision signals. Guler et al.\ \cite{guler2018densepose} published a method to regress human surface coordinates using a Mask-RCNN architecture. They also released the COCO (Common Objects In Context) Denspose dataset that consists of human 2D images annotated with human semantic part segmentation and UV coordinate mapping within each semantic part \cite{guler2018densepose}. These UV coordinate mappings establish human correspondence with the 3D human model \cite{loper2015smpl}. To create the annotations, humans were hired who spent a lot of time to create annotations that where rather sparse and prone to errors. This is a big limitation for a task like dense correspondence where the reliability on human annotators results in the creation of erroneous data. Our work focuses on resolving this issue with the dataset. 


Our contributions are two fold. First, we present a method to automatically generate synthetic data and extract ground truth annotations for the problem of dense 2D-3D human correspondence. The data generated consists of human semantic segmentation, exact 2D and 3D human pose keypoints and ground truth 2D pixels to 3D human voxel mappings. Second, we evaluate the performance of our labelled synthetic data for the problem of dense correspondence mapping between 2D human images and 3D human SMPL models \cite{loper2015smpl} using the method in \cite{guler2018densepose}. 


\vspace{-2mm}

\section{Methodology}
\vspace{-2mm}
In this section, we describe the completely automated method used to generate accurate ground truth UV data of a human mesh using a synthetic 3D environment where human avatars are made to perform human actions using motion capture (MOCAP) data.
\vspace{-2mm}

\subsection{Avatar animation and rendering}
\label{Synthetic data generation}
\vspace{-1mm}

Part of the MOCAP data was collected at the Queensland Centre for Advanced Technologies. Ten volunteers performed a set of actions (e.g., walking, running, seating) while wearing reflective markers. The 3D position of these markers was captured using 24 synchronised infrared cameras and Qualisys Track Manager was used to reconstruct and animate 3D human skeletons. Additional publicly available MOCAP data were included to increase diversity \cite{yang2019open}.

The SMPL parametric model was used to create the human avatar \cite{loper2015smpl}. It is a compact representation of the human body as an average human surface paired with modes of variation (statistical deformations driving vertices).  It uses user-defined shape known as blend-shapes to create the general shape of the human avatar, and pose blend-shapes that are automatically updated by the model based on the human pose to account for tissue deformation. The model can thus be used to generate an infinite number of statistically plausible human body shapes, which animate realistically based on MOCAP data \cite{loper2015smpl}.


Publicly available textures from the Surreal dataset\cite{corl2018surreal} were mapped onto the avatar to generate a realistic human look. The male and female textures were separated and the semantic parts (torso, arms etc.) mixed and stitched together to increase variety. To render realistic human images, we used 3456 random background images from the COCO Densepose train set and the synthetic human model is made to move in 3D space between the cone of the camera and image plane based on animation. 

The Unity Real-Time Development Platform was used to render all video data. 9 different synthetic cameras were used to create synthetic data where avatar animation is overlayed onto the existing background image that is placed within the cone of the rendering camera. Synthetic camera noise and camera lens distortion is added to the rendered scene to impart more realism. Separate SMPL human models are used for male and female bodies to accommodate precise physiological representation. We used a single windows desktop with GPU to render the images.




\begin{figure}
     \centering
        \includegraphics[width=0.9\linewidth]{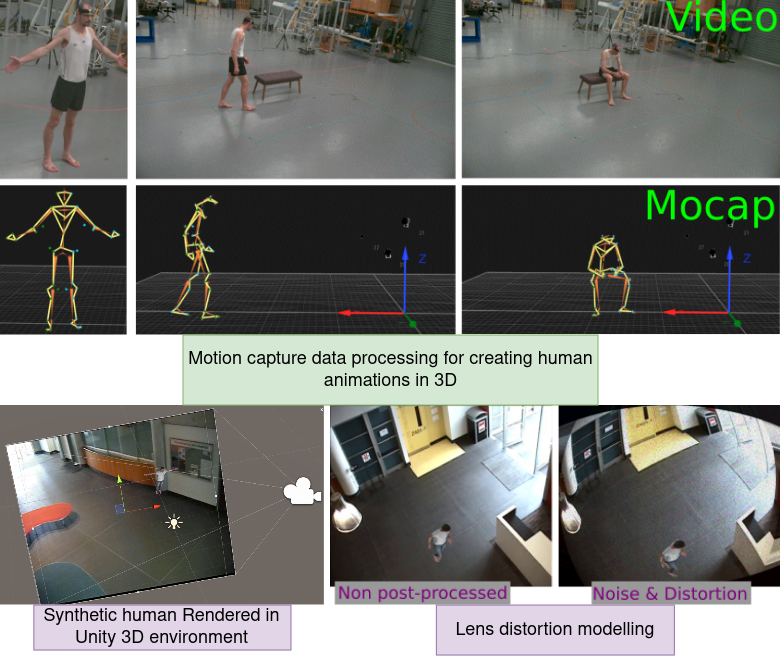}
        \vspace{-2mm}
        \caption{Synthetic Data Generation Pipeline}
        \label{fig:Datagen_pipeline}
        \vspace{-3mm}
\end{figure}




\subsection{Mesh Vertex visibility test in 3D}
\label{Mesh Vertex visibility test in 3D}
The vertex visibility in 3D is dependent on the camera view $\alpha$ and pose parameters $\theta$. We compute the concave hull of the SMPL human mesh in 3D from the mesh vertices to get precise visible vertex locations of the mesh surface. We used the method of Ray casting from each vertex locations to the corresponding camera center to solve the vertex visibility for a given view.  All visible vertices in the world coordinate 3D space are transformed into image space in-order to retrieve visible vertices of the human 3D mesh in the rendered 2D image space. 


\begin{equation}\label{eqn_1}
    Visibility(v_i) = F(\alpha, \theta),
\end{equation}
where $v_i$ denotes 3D vertex coordinate and $\alpha$, $\theta$ are camera parameters, pose parameters respectively.

\subsection{Semantic region mapping and IUV segmentation}\label{Semantic region mapping and IUV segmentation}
\begin{figure}
        \includegraphics[width=\linewidth]{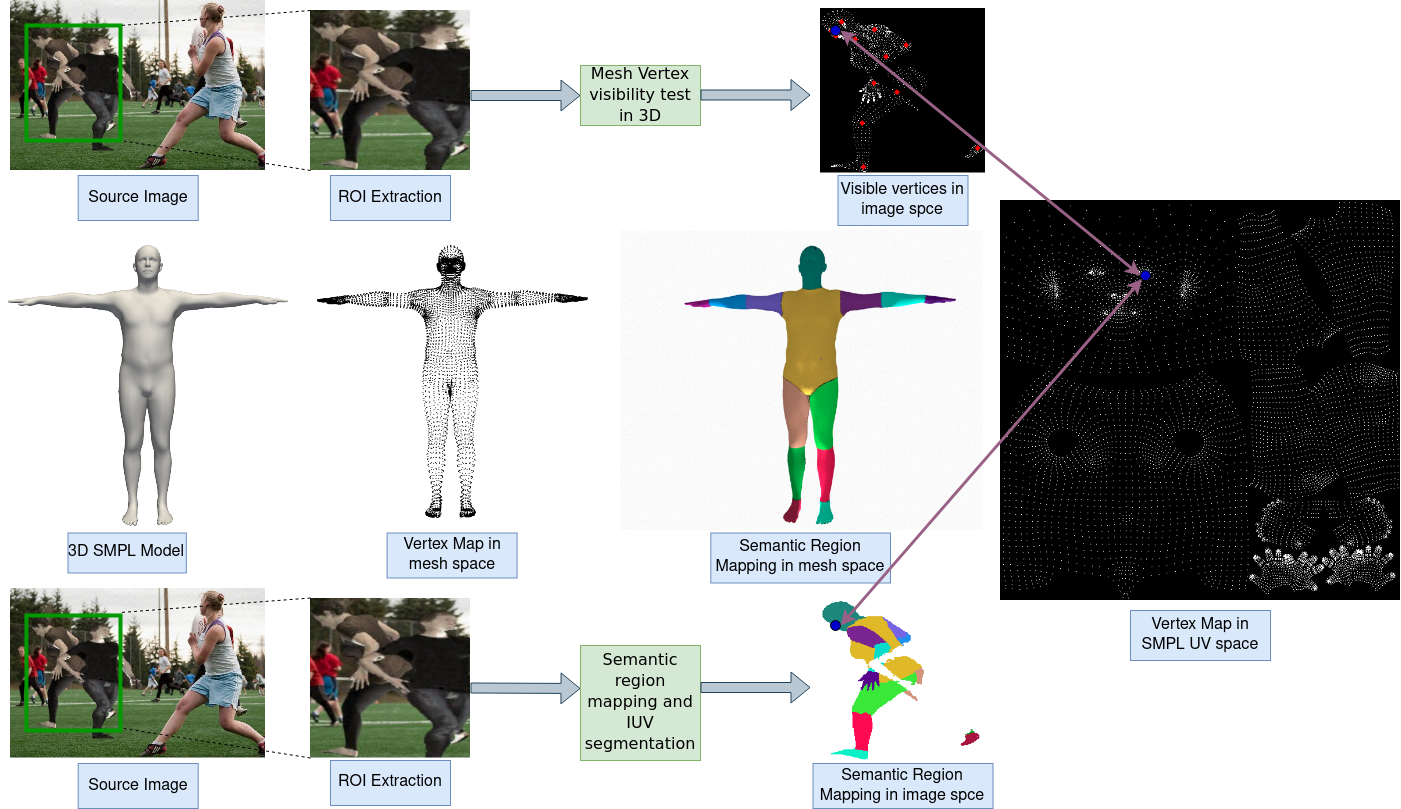}
        \caption{Human Dense Correspondence Mapping on SMPL Human model}
        \label{DP_mapping}
\end{figure}

We divide the human mesh in SMPL UV space into 24 part atlas UV space as discussed in \cite{guler2018densepose}. The whole human mesh is divided into 14 semantic regions like Head, Torso, Lower/Upper Arms, Lower/Upper Legs, Hands and Feet etc. To extract ground truth segmentations, all image space coordinates are projected on to the world coordinate space to see if these points collide with the concave hull of the human mesh. All such coordinates of collision points in 2D are obtained as a dense mapping of the human mesh in the image space. To extract semantic region mapping, we project all image space coordinates from the corresponding camera to check whether the rays hit the concave hull field. If a point in image space when casted, hits the field, then the collision point is associated with the geometrically closest vertex coordinate in world coordinate space such that a correlation is established between all 2D points in image space on the human surface and mesh vertices. The association created between image space coordinates and UV space coordinates helps to gather precise semantic region information in the image domain as shown in Figure \ref{DP_mapping}.








\subsection{Occlusion generation and Alpha blending}
\vspace{-1mm}
\label{Occlusion generation and Alpha blending of occluded objects }
To create occlusions, we used 227 segmented 2D mundane objects which were extracted using a background removal tool. An object is randomly sampled and re-scaled according to the human Region of Interest (ROI) in the scene. Then, the object is positioned randomly over the the rendered human in the image. The 2D objects that are used to create occlusion have very sharp edges when randomly positioned on top of the human surface. To provide smooth transition between the edge of the 2D object and the rendered image , a random Gaussian filter and alpha blending are applied over a thick inner and outer boundary of the object. This allows for seamless transition between object edge and background.

\subsection{Scene harmonisation for visual data generation}
\begin{figure}
\label{Scn_harmo}
    \includegraphics[width=\linewidth]{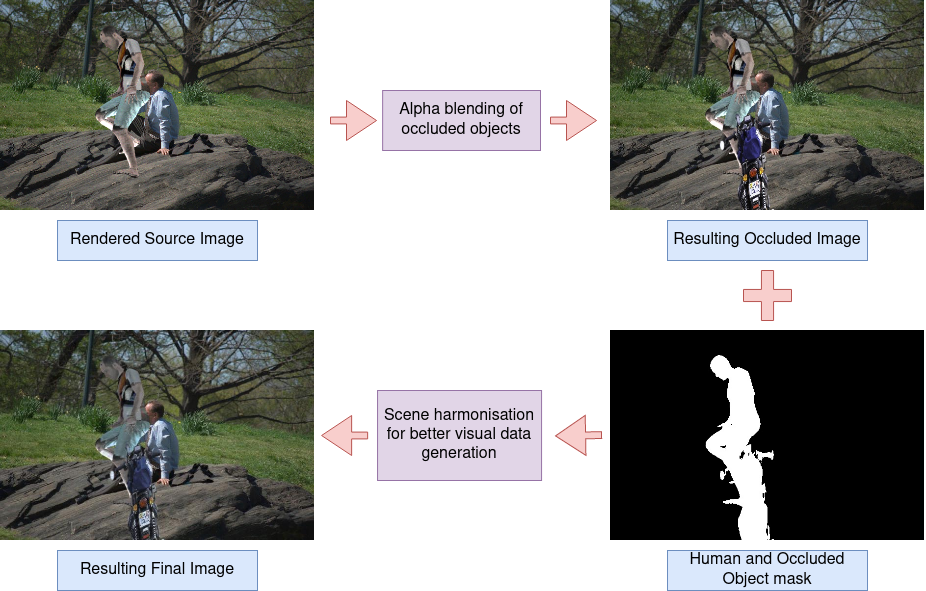}
    \caption{Synthetic Human Dataset Post Processing}
    \label{fig:scn_harmo}
\end{figure}

\vspace{-1mm}
\label{Scene harmonisation for better visual data generation}
To better match the ``style'' of the human avatar and occluding object with the background, a scene harmonisation method was used \cite{ling2021region}. It uses a  generative adversarial network conditioned on the ROI (the human mask and the occluded object mask) that transforms the target human pixel features to adapt more with the background style of the image as depicted in Figure \ref{fig:scn_harmo}. 

\section{EXPERIMENTS \& RESULTS}
\vspace{-2mm}
\label{sec:Experiments}
\subsection{COCO Densepose Training}
\label{COCO Densepose Training}
\vspace{-1mm}
 Our goal is to evaluate  the performance of synthetic data generated for the problem of dense correspondence estimation. We used the detectron2 for training the dense correspondence model. For all experiments, we used Resnet-50 backbone as the primary feature extractor and the Panoptic head from \cite{kirillov2019panoptic} and DeepLabV3 head from \cite{chen2017rethinking}. A hyperparameter free training strategy is followed according to the linear scaling rule from \cite{goyal2017accurate}. Comparison between the same model setting trained using real data and synthetic data are discussed in the coming sub-sections. We discovered that training the model from scratch using our synthetic data caused the model to over-fit the synthetic data in fewer iterations, thus losing generalisation capability on real data. To avoid this, we initialised the model with weights trained on real data and only the last layers i.e, the part of the model where the IUV estimation occurs, is fine-tuned. We also used an additional mask-based supervision for training the coarse segmentation parts (14 parts) whose weights are randomly initialised. Thus, the human mask estimation and the IUV dense correspondence estimation is completely learned from our synthetic data. We trained the model for 35,000 iterations with batch size 2, which is almost equivalent to an epoch at a learning rate 0.001 on a single Nvidia Tesla T4 GPU to avoid over-fitting. To train the model, we used 73,671 synthetic human instances containing directly rendered images without occlusion and images post-processed by adding occlusion along with scene harmonisation.

 \begin{table}
\resizebox{\linewidth}{!}{
\begin{tabular}{ |c|c|c|c|c|c|c| } 
\hline
Task & \textbf{AP} & $\textbf{AP}_{50}$ & $\textbf{AP}_{75}$ &  \textbf{AR} & $\textbf{AR}_{50}$ & $\textbf{AR}_{75}$  \\ \hline
\hline
\multirow{1}{*}{bbox} & 58.35 & 84.89 & 63.99 &  &  &  \\ \hline
\multirow{1}{*}{GPS} & 52.18 & 85.69 & 57.96 & 61.07 & 91.48 & 69.19 \\ \hline

\multirow{1}{*}{GPSm} & 54.38 & 88.02 & 63.79 & 62.08 & 93.40 & 72.89 \\ \hline

\multirow{1}{*}{Segm} & 57.30 & 91.69 & 66.62 & 64.91 & 94.07 & 76.19 \\ \hline
\hline
\end{tabular}
}
\vspace{-2mm}
\caption{\label{tab:tab-1}Model Fine-tuned with Synthetic data}
\end{table}

\begin{table}
\resizebox{\linewidth}{!}{
\begin{tabular}{ |c|c|c|c|c|c|c| } 
\hline
Task & \textbf{AP} & $ \textbf{AP}_{50}$ & $ \textbf{AP}_{75}$ &  \textbf{AR} & $\textbf{AR}_{50}$ & $\textbf{AR}_{75}$  \\ \hline
\hline
\multirow{1}{*}{bbox} & 61.08 & 88.04 & 67.10 &  &  &  \\ \hline
\multirow{1}{*}{GPS} & 65.54 & 91.73 & 75.36 & 73.02 & 95.09 & 81.32 \\ \hline
\multirow{1}{*}{GPSm} & 66.69 & 92.74 & 79.62 & 72.61 & 96.03 & 84.31 \\ \hline

\multirow{1}{*}{Segm} & 68.33 & 94.54 & 81.51 & 73.74 & 96.70 & 86.22 \\ \hline
\hline
\end{tabular}
}
\vspace{-2mm}
\caption{\label{tab:tab-2}
Model trained on real data (Pre-trained model available in detectron2 repo)}
\end{table}

\begin{table}
\resizebox{\linewidth}{!}{
\begin{tabular}{ |c|c|c|c|c|c|c| } 
\hline
Task & \textbf{AP} & $ \textbf{AP}_{50}$ & $ \textbf{AP}_{75}$ &  \textbf{AR} & $\textbf{AR}_{50}$ & $\textbf{AR}_{75}$  \\ \hline
\hline
\multirow{1}{*}{bbox} & 61.08 & 88.04 & 67.10 &  &  &  \\ \hline
\multirow{1}{*}{GPS} & 53.87 & 88.18 & 59.54 & 61.25 & 91.84 & 68.83 \\ \hline

\multirow{1}{*}{GPSm} & 56.56 & 90.52 & 66.90 & 62.44 & 93.89 & 73.25 \\ \hline

\multirow{1}{*}{Segm} & 59.07 & 92.44 & 69.67 & 65.38 & 94.78 & 76.59 \\ \hline
\hline
\end{tabular}
}
\vspace{-2mm}
\caption{\label{tab:tab-3}
Model Fine-tuned with Synthetic data inferred with real bbox}
\end{table}

\begin{figure}
\label{Mask_Comp}
\centering
    \includegraphics[width=0.9\linewidth]{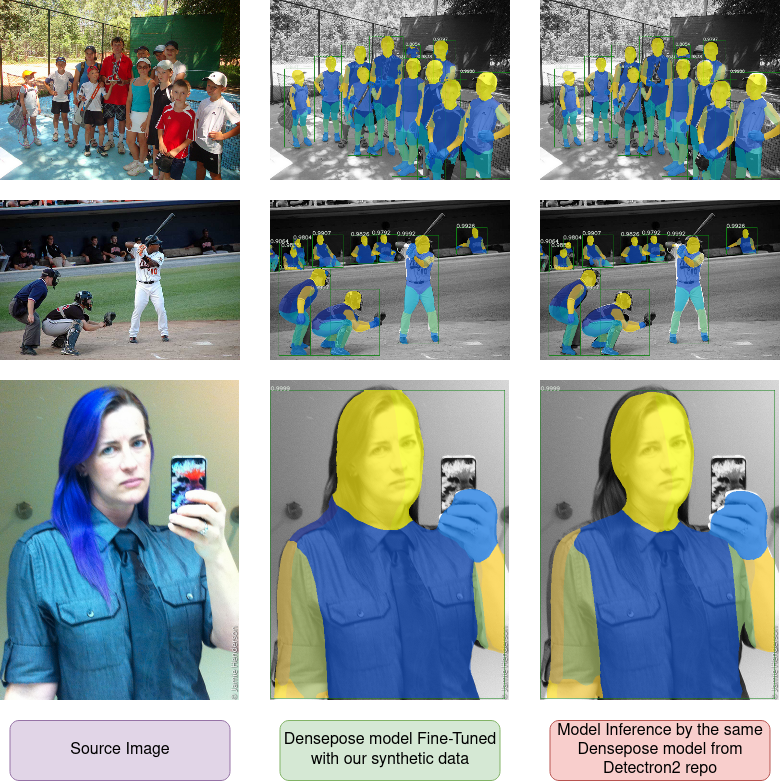}
    \vspace{-2mm}
    \caption{Semantic part segmentation comparison }
    \label{fig:Mask_comp}
    \vspace{-2mm}
\end{figure}

\begin{figure}
\label{Tex_Comp}
\centering
    \includegraphics[width=0.9\linewidth]{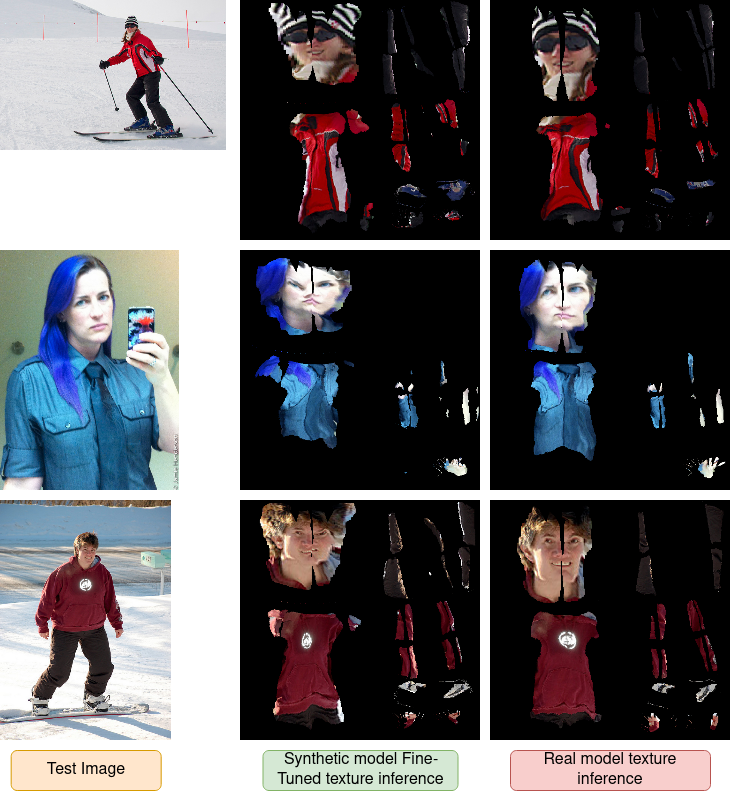}
        \vspace{-2mm}
    \caption{Texture extraction comparison}
    \label{fig:Tex_comp}
    \vspace{-2mm}
\end{figure}

\subsection{Evaluation Metrics}
\vspace{-1mm}
\label{Evaluation Metrics}
We report the standard COCO densepose evaluation metrics described in \cite{guler2018densepose}. We consider the human bounding box estimation (bbox), human segmentation mask estimation (Segm), Geodesic Point Similarity(GPS) and Geodesic Point Similarity within the correctly predicted human part mask (GPSm) to evaluate the dense correspondence. For all tasks, \textbf{mIoU} (mean Intersection over Union) is considered with thresholds and \textbf{AP} (average precision averaged over IoU thresholds) is the primary metric followed by $ \textbf{AP}_{50}$ and $\textbf{AP}_{75}$ being the selected supplementary metrics.

\subsection{Quantitative Evaluation}
\vspace{-1mm}
\label{Quantitative Evaluation}
The COCO densepose test set containing 1508 images with 5581 person instances are used for performing quantitative evaluation. Table \ref{tab:tab-1} shows the results of the dense correspondence model fine-tuned with synthetic data. Table \ref{tab:tab-2} shows the state-of-the-art results\cite{guler2018densepose}. Table \ref{tab:tab-3} shows another experiment where the bounding box estimation is performed by the state-of-the-art model while all the remaining metrics are evaluated by the model fine-tuned with synthetic data. The results depict that with better known human ROIs, all other metrics showcase improvements.

\subsection{Qualitative Evaluation}
\label{Qualitative Evaluation}
We provide some comparative results between the human semantic region estimation by the model fine-tuned using synthetic data and estimation performed by the pre-trained model trained on real data in Figure \ref{fig:Mask_comp}. The human pixels inferred within the ROI are then transformed to the SMPL UV space for further comparison of the 2D human image to 3D surface mapping. Figure \ref{fig:Tex_comp} shows comparison of the human textures inferred by the model fine-tuned with synthetic data and the pre-trained model trained using real data.

\section{CONCLUSION}
\label{sec:Conclusion}
We studied the possibility of training deep-learning models for human 2D-3D mapping using synthetic data, which has the advantage of being dense, precise, unbiased, labour free and solves issues of data volume and privacy. Initial experiments show promising quantitative results and qualitative results that the synthetic model has the potential to improve fine edge accuracy. The current gap between synthetic and real data can cause an over-fitting of the model to the synthetic data, which can be solved using advanced domain adaptation techniques. This presents a clear avenue for work on synthetic data for 2D-3D mapping.

\vfill\pagebreak


\bibliographystyle{IEEEbib}
\bibliography{strings,refs}

\end{document}